\UseRawInputEncoding
\documentclass{article}

\usepackage[preprint]{neurips_2022}

\usepackage[utf8]{inputenc} 
\usepackage[T1]{fontenc}    
\usepackage{hyperref}       
\usepackage{url}            
\usepackage{booktabs}       
\usepackage{amsfonts}       
\usepackage{nicefrac}       
\usepackage{microtype}      
\usepackage{xcolor}         
\usepackage{graphicx}
\usepackage{float}
\usepackage{algorithm}
\usepackage{algorithmic}
\usepackage{makecell}
\usepackage{subfigure}
\usepackage{natbib}
\setcitestyle{numbers,square}
\usepackage{amsmath,amsfonts,amssymb,amsthm, bm}
\title{ABS-SGD: A Delayed Synchronous Stochastic Gradient Descent Algorithm with Adaptive Batch Size for Heterogeneous GPU Clusters}

\author{%
  Xin Zhou\qquad Ling Chen\thanks{Correspondence to lingchen@cs.zju.edu.cn}
\qquad Houming Wu \\
College of Computer Science\\
Zhejiang University\\
}

\begin{document}

\maketitle

\begin{abstract}
As the size of models and datasets grows, it has become increasingly common to train models in parallel. However, existing distributed stochastic gradient descent (SGD) algorithms suffer from insufficient utilization of computational resources and poor convergence in heterogeneous clusters. In this paper, we propose a delayed synchronous SGD algorithm with adaptive batch size (ABS-SGD) for heterogeneous GPU clusters. In ABS-SGD, workers perform global synchronization to accumulate delayed gradients and use the accumulated delayed gradients to update parameters. While workers are performing global synchronization for delayed gradients, they perform the computation of the next batch without specifying batch size in advance, which lasts until the next global synchronization starts, realizing the full utilization of computational resources. Since the gradient delay is only one iteration, the stale gradient problem can be alleviated. We theoretically prove the convergence of ABS-SGD in heterogeneous clusters. Extensive experiments in three types of heterogeneous clusters demonstrate that ABS-SGD can make full use of computational resources and accelerate model convergence: When training ResNet18 network with 4 workers, ABS-SGD increases the convergence speed by 1.30× on average compared with the best baseline algorithm.
\end{abstract}

\section{Introduction}
The performance of deep learning models is positively correlated with the size of models and datasets. As the size of models and datasets grows, due to the limitation of single machine memory and the requirement for training speed, it has become increasingly common to train models in parallel. Distributed data parallel training replicates the model to multiple workers and workers communicate with each other according to specific distributed stochastic gradient descent (SGD) algorithms, which has become a dominant strategy due to its minimal invasiveness.

Distributed data parallel training is typically applied in homogeneous clusters with similar worker performance \cite{goyal2017accurate,jia2018highly}, but it is also often applied in heterogeneous clusters with heterogeneous hardware performance or shared by multiple users \cite{hazelwood2018applied,jeon2018multi}. The heterogeneity in clusters can be static or dynamic. Static heterogeneity refers to the differences in hardware performance between workers, e.g., different CPU and GPU configurations. Dynamic heterogeneity is due to resource competition or transient performance fluctuations, regardless of the presence of static heterogeneity in a cluster.

Early distributed SGD algorithms assume models are trained in homogeneous clusters. Synchronous SGD algorithms, e.g., bulk synchronous parallel SGD (BSP-SGD) \cite{gerbessiotis1994direct}, require synchronization between workers to accumulate gradients. The iteration time is affected by slow workers in heterogeneous clusters, which is called the straggler problem \cite{cipar2013solving,zinkevich2010parallelized}. Asynchronous SGD algorithms, e.g., asynchronous parallel SGD (ASP-SGD) \cite{lian2015asynchronous}, do not require synchronization between workers and fast workers can iterate more times than slow workers in heterogeneous clusters, but the gradients used to update the global model's current parameters may be too stale for the slow workers and impart model convergence, which is called the stale gradient problem \cite{ho2013more}.

To address these issues, many distributed SGD algorithms for heterogeneous clusters have been proposed. Some works alleviate the straggler problem by redundant execution, which synchronize gradients between a certain proportion of workers according to their iteration speed. Since slower workers serve as redundant workers and do not participate in synchronization, these algorithms waste the computational resources of slower workers. Some works are proposed to balance the load of workers by performance prediction, which predict worker performance and allocate corresponding loads, e.g., batch size. However, in dynamic heterogeneous clusters, there are many factors that affect the performance of workers, which cannot be completely modelled by a performance predictor, leading to large prediction errors and insufficient utilization of computational resources. Some works alleviate the stale gradient problem by controlling the staleness of delayed gradients, e.g., limiting the difference between the iteration numbers of fast and slow workers. These algorithms are applicable to clusters with only dynamic heterogeneity, where the performance degradation of slow workers is temporary and slow workers can catch up with fast workers quickly. However, in static heterogeneous clusters, fast workers always have to wait for slow workers, leading to insufficient utilization of computational resources.

To address the aforementioned problems, we propose delayed synchronous stochastic gradient descent algorithm with adaptive batch size (ABS-SGD) for heterogeneous GPU clusters.  The algorithm uses the heterogeneous-aware delayed synchronous parallel mechanism to achieve adaptive batch size, which uses delayed gradients to update parameters, and performs global synchronization between delayed gradients. Each worker performs the computation of the next batch without specifying batch size in parallel with the global synchronization for delayed gradients. Workers keep computing until the next global synchronization starts, which makes full use of computational resources. In addition, since the gradient delay is only one iteration, the stale gradient problem can be alleviated. On this basis, the weighted gradient averaging mechanism is introduced to address the inconsistent batch size of each worker, and the delayed gradient compensation mechanism is introduced to improve the quality of delayed gradients to ensure high-quality parameter update.

The contributions of this paper are summarized as follows:
\begin{itemize}
    \item We propose ABS-SGD, which mainly includes the following three mechanisms: (1) heterogeneous-aware delayed synchronous parallel mechanism to achieve adaptive batch size; (2) weighted gradient averaging mechanism; (3) delayed gradient compensation mechanism to ensure high-quality parameter update in heterogeneous clusters.
    \item From the theoretical perspective, we prove that ABS-SGD has a linear iteration speedup with respect to the total computational resources of a heterogeneous cluster.
    \item We conduct extensive comparative experiments in three types of heterogeneous GPU clusters, and the experimental results show that ABS-SGD achieves a time speedup by 1.30× on average compared with the best baseline algorithm.
\end{itemize}

\section{Related work}
Many distributed SGD algorithms for heterogeneous clusters have been proposed to solve the straggler and stale gradient problems of traditional distributed SGD algorithms. The ideas of these algorithms are mainly based on redundant execution, delayed gradient control, and performance prediction.

Redundant execution strategies \cite{ananthanarayanan2013effective,zaharia2008improving} are widely applied in traditional data analytics frameworks, in which multiple replicates of a task are initiated and only the result from the first completed task is accepted. This idea has also been introduced to the field of distributed machine learning. Chen et al. \cite{chen2016revisiting} propose to train deep learning models with additional backup workers, using the gradients from the earliest completed portion of workers. This algorithm is implemented based on parameter server. Li et al. \cite{li2020taming} propose a series of partial collective operations based on allreduce to coordinate synchronization between workers. However, redundant execution is merely a suboptimal solution, as it may still hinder overall training speed if the proportion of backup workers is not set properly, and it fails to utilize the computational resources of the backup workers.

The distributed SGD algorithms based on delayed gradient control prevent the model convergence from being affected by delayed gradients, which is realized by controlling the staleness of delayed gradients when the model is trained asynchronously. Starting from ASP-SGD, SSP-SGD \cite{ho2013more} stipulates that the difference in the iteration numbers between the fast and slow workers needs to be smaller than a given staleness threshold, so as to avoid delayed gradients being too stale to keep model convergence. Based on SSP-SGD, Jiang et al. \cite{jiang2017heterogeneity} improve model convergence speed in heterogeneous clusters by adjusting the learning rate, i.e., giving a lower learning rate to the gradient that is too stale. Since fast workers are always faster than slow workers in static heterogeneous clusters, fast workers always have to wait for slow workers. The above algorithms still cannot make full use of computational resources in static heterogeneous clusters. To solve this problem, Zhao et al. \cite{zhao2019dynamic} reduce the waiting time of fast workers by dynamically adjusting the staleness threshold and the adjustment of the staleness threshold is related to the performance gap between workers. However, when the performance gap between workers is too large, the staleness threshold also needs to be adjusted greatly, and the stale gradient problem still exists.

The algorithms based on performance prediction assign appropriate load to each worker by predicting worker performance, which can be divided into synchronous and asynchronous algorithms. BSP-SGD and Local SGD \cite{stich2018local} are classical synchronous SGD algorithms. Based on BSP-SGD, some works propose to allocate batch size by the predicted worker performance. Ye et al. \cite{ye2020dbs} assume that worker performance will not change significantly in the short term and use the most recent computation time of workers as a measure of performance to allocate batch size in the next iteration. This algorithm does not consider long-term computation information and can only make rough predictions of worker performance, so Tyagi et al. \cite{tyagi2020taming} propose using exponential averages to predict worker performance. Chen et al. \cite{chen2020semi} consider the non-proportional relationship between computation time and batch size in GPU clusters and propose different performance prediction strategies for CPU and GPU clusters. For CPU clusters, a nonlinear autoregressive with exogenous inputs (NARX) model is used for prediction. For GPU clusters, a numerical approximation method is used, in which the batch sizes of the fastest and slowest workers are dynamically adjusted according to the computation time at each iteration. Based on Local SGD, Stripelis et al. \cite{stripelis2021semi} and Cao et al. \cite{cao2021sap} propose that workers perform different local iteration numbers on their local models according to worker performance, and then the workers take a weighted average of their local models according to their respective local iteration numbers to get the global model.

There are also some works based on asynchronous algorithms, which propose dividing workers based on worker performance before training. Jiang et al. \cite{jiang2019novel} use K-means clustering to achieve grouping and synchronize gradients within groups while training asynchronously between groups. Sun et al. \cite{sun2021gssp} use the Jenks Natural Breaks algorithm for grouping and train asynchronously within groups while periodically synchronizing the model between groups. In dynamic heterogeneous clusters, due to the large number of factors that can dynamically affect worker performance and the inability of performance predictors to model all of these factors, these algorithms based on performance prediction have large prediction errors and fail to fully utilize the computational resources of heterogeneous clusters.

\section{Preliminary}
\subsection{Problem definition}
Train a model with $n$ parameters can be formulated as a non-convex optimization problem:
\begin{equation}
\min_{x \in R^{n}}{f(x) = E_{\xi}\left\lbrack {F\left( {x;\xi} \right)} \right\rbrack},
\end{equation}
where $\xi \in \Xi$ is a random variable and $f(x)$ is a smooth function. The most common specification is that $\Xi$ is an index set of all training samples $\Xi = \left\{ 1,2,\ldots,D \right\}$ and $F(x;\xi)$ is the loss function with respect to the training sample indexed by $\xi$.
\subsection{Notations}
\begin{itemize}
\item $\left\| \cdot \right\|$ indicates the L2 norm of a vector.
\item 	$< \cdot , \cdot >$ indicates the inner product of two vectors $x$ and $y$.
\item $g(x;\xi)$ is used to denote $\nabla f(x;\xi)$ for short.
\item $E( \cdot )$ means taking the expectation in terms of all random variables.
\end{itemize}

\section{Assumptions}
Throughout this paper, we make the following assumptions for the objective function. All of them are quite common in the analysis of SGD algorithms.

\textbf{Assumption 1} (Lipschitzian Gradient) The objective function $f(x)$ is differentiable, and its gradient is L-Lipschitz continuous:
\begin{equation}
\left\| {f(x) - f(y)} \right\| \leq L\left\| {x - y} \right\|,L \geq 0.
\end{equation}
There is a common corollary to this assumption:
\begin{equation}
\left| {f(x) - f(y) - \left\langle {\nabla f(y),x - y} \right\rangle} \right| \leq \frac{L}{2}\left\| {x - y} \right\|^{2}.
\end{equation}

\textbf{Assumption 2} (Unbiased Gradient) The stochastic gradient $g\left( {x;\xi} \right)$ is unbiased:
\begin{equation}
\nabla f(x) = E\left\lbrack {g\left( {x;\xi} \right)} \right\rbrack.
\end{equation}

\textbf{Assumption 3} (Unbiased Gradient) The variance of stochastic gradient $g\left( {x;\xi} \right)$ is bounded:
\begin{equation}
E\left\lbrack \left\| {g\left( {x;\xi} \right) - \nabla f(x)} \right\|^{2} \right\rbrack \leq \sigma^{2}.
\end{equation}
\begin{figure}[H]
  \centering
  \includegraphics[width=12cm,height=7cm]{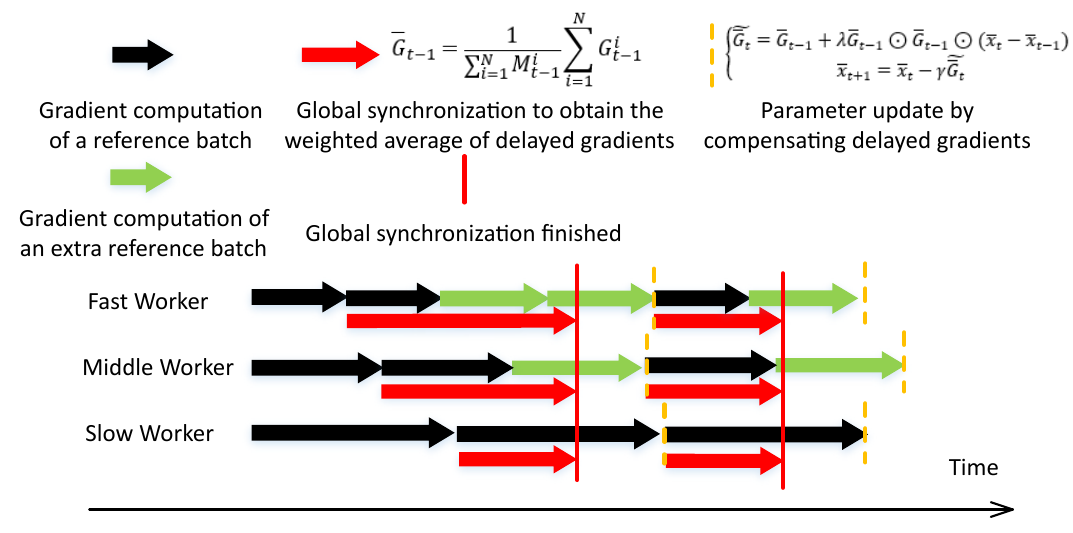}
  \caption{The training process of ABS-SGD (best viewed in color). The formulas on the figure represent the specific meanings of operations, and details are given in Algorithm \ref{algorithm1}.}
  \label{process_fig}
\end{figure}

\section{Algorithm}
The training process of ABS-SGD is shown in Figure \ref{process_fig}. ABS-SGD mainly includes the following three mechanisms: (1) heterogeneous-aware delayed synchronous parallel mechanism; (2) weighted gradient averaging mechanism; (3) delayed gradient compensation mechanism.

\begin{algorithm}[H]
    \caption{ABS-SGD}
    \label{algorithm1}
    \textbf{Input}: The number of workers $N$, the number of iterations $T$, reference batch size $M_r$, learning rate $\gamma$, initial parameters $\overline{x}_0$, compensation coefficient $\lambda$. \\
    \textbf{Output}: Optimization result $\overline{x}$.
    \begin{algorithmic}[1] 
        \STATE Initialization: $t = 0,G_{- 1}^{i} = 0,i = 1,2,\ldots,N$.
        \WHILE{$t<T$}
        \STATE Run Step 1 and Step 2 in parallel in worker $W_i$:
        \STATE Step 1: \\
        \qquad Obtain the weighted average of delayed gradients by global synchronization:\\
        \qquad ${\overline{G}}_{t - 1} = \frac{1}{\sum\limits_{i = 1}^{N}M_{t - 1}^{i}}{\sum\limits_{i = 1}^{N}G_{t - 1}^{i}}$.\\
        \STATE Step 2: \\
        \qquad Initialization: \\
        \qquad \qquad The accumulated gradients of worker $W_i$ at $t$-th iteration: $G_t^i=0$. \\
        \qquad \qquad The accumulated batch size of worker $W_i$ at $t$-th iteration: $M_t^i=0$.\\
        \qquad \textbf{while} Step 1 not finished  \textbf{do} \\
        \qquad \qquad Compute and accumulate the gradients of reference batch: \\
        \qquad \qquad $G_{t}^{i} = G_{t}^{i} + {\sum\limits_{j = M_{t}^{i}}^{M_{t}^{i} + M_{r}}{\nabla f_{i}\left( {{\overline{x}}_{t};\xi_{t,j}^{i}} \right)}}$.\\
        \qquad \qquad Obtain actual batch size: $M_{t}^{i} = M_{t}^{i} + M_{r}$.\\
        \qquad \textbf{end while} \\
        \qquad Compensate delayed gradients: $\widetilde{{\overline{G}}_{t}} = {\overline{G}}_{t - 1} + \lambda{\overline{G}}_{t - 1} \odot {\overline{G}}_{t - 1} \odot \left( {\overline{x}}_{t} - {\overline{x}}_{t - 1} \right)$. \\
        \qquad Update parameters: ${\overline{x}}_{t + 1} = {\overline{x}}_{t} - \gamma\widetilde{{\overline{G}}_{t}}$.\\
        \STATE $t=t+1$.
        \ENDWHILE
        \STATE \textbf{return} $\overline{x} = {\overline{x}}_{T}$.
    \end{algorithmic}
\end{algorithm}

In order to fully utilize computational resources and alleviate the stale gradient problem, we propose the heterogeneous-aware delayed synchronous parallel mechanism, so that the batch size of each worker in each iteration can be adapted to its performance.
\begin{itemize}
\item Heterogeneous-aware delayed synchronous parallel mechanism: Traditional synchronous SGD algorithms use the gradients ${\overline{G}}_{t}$ when updating the parameters ${\overline{x}}_{t}$, but in this parallel mechanism, the delayed gradients ${\overline{G}}_{t-1}$ are used. The computation of the current gradients $G_t^i$ are parallel with the global synchronization of the delayed gradients ${\overline{G}}_{t-1}$. For the $t$-th iteration, when the model parameters are ${\overline{x}}_{t}$, each worker simultaneously computes the current gradients $G_t^i$ corresponding to ${\overline{x}}_{t}$ and obtains the delayed gradients ${\overline{G}}_{t-1}$ by the global synchronization, and the computation will continue until the global synchronization completes. In order to ensure that the computation of the current gradients can continue, gradients of multiple batches may be computed in an iteration. We call each batch in an iteration as reference batch with corresponding reference batch size $M_r$. Workers will perceive the synchronization state of delayed gradients after completing the gradient computation of a reference batch. If the synchronization is not finished, workers will continue to compute the current gradients of the next reference batch and accumulate gradients between reference batches. If finished, workers will stop the computation. Therefore, if a worker computes gradients of $k$ reference batches in an iteration, the batch size of the worker in this iteration is $kM_r$. The batch size of this iteration is the accumulated batch size of all workers.
\end{itemize}
On the other hand, we introduce the following two mechanisms to solve the problem of inconsistency in the batch size of each worker and improve the quality of delayed gradients:

\begin{itemize}

\item Weighted gradient averaging mechanism: After introducing the heterogeneous-aware delayed synchronous parallel mechanism, the batch size of each worker is adapted to its performance. Directly averaging the gradients computed by each worker would cause statistical deviation, so we perform a weighted average on the gradients based on the respective batch size of each worker. In the implementation, each worker synchronizes their batch size before synchronizing their gradients in each iteration, so that each worker can know the total batch size of the iteration. 
\item Delayed gradient compensation mechanism: The heterogeneous-aware delayed synchronous parallel mechanism uses delayed gradients to update parameters, and the gradient delay is fixed to one iteration. Therefore, we introduce the delayed gradient compensation mechanism based on first-order Taylor expansion \cite{zheng2017asynchronous}, so that delayed gradients can approach the current gradients to improve the update quality. The formula for compensating $G_{t-1}$ to $\widetilde{G_t}$ is: \begin{equation}
    \widetilde{G_{t}} = G_{t - 1} + \lambda G_{t - 1} \odot G_{t - 1} \odot \left( {x_{t} - x_{t - 1}} \right),
\end{equation}
 where $\odot$ represents element-wise multiplication, and $\lambda$ is the compensation coefficient that needs to be adjusted. This formula has a theoretical basis and is computationally efficient. Since the compensation of the delayed gradients $G_{t-1}$ requires the corresponding delayed parameters $x_{t-1}$, this mechanism introduces some memory overhead.
\end{itemize}
Algorithm \ref{algorithm1} summarizes the complete training process of ABS-SGD. Each iteration contains two steps that are parallel with each other. In Step 1, global synchronization is conducted to obtain the weighted average of delayed gradients. In Step 2, the current gradients are continuously computed until step 1 is finished. Then the delayed gradients obtained by synchronization are compensated, and model parameters are updated.

\section{Theoretical analysis}
To analyze Algorithm 1, besides Assumptions 1-3, we make the following additional assumptions.

\textbf{Assumption 4} (Independent Sample) All random variables $\xi$ representing the sample are independent to each other.

\textbf{Assumption 5} (Bounded Delay) The batch size for each iteration is bounded:
\begin{equation}
NM_{r} \leq M_{t} \leq KM_{r}.
\end{equation}
In non-convex optimization, the weighted average of the L2 norm of the gradient in $T$ iterations is used as the criterion for evaluating the convergence rate.

\textbf{Theorem 1} Assume that Assumptions 1-5 hold and the learning rate $\gamma_{t}$ in Algorithm 1 satisfies: \\
\begin{equation}
{\gamma_{t}}^{2}L{M_{t}}^{2} + L\gamma_{t}M_{t} \leq 1.
\end{equation}
We have the following ergodic convergence rate for the iteration of Algorithm 1:
\begin{equation}
\frac{1}{\sum\limits_{t = 1}^{T}\gamma_{t}}{\sum\limits_{t = 1}^{T}{\gamma_{t}\left\| {\nabla f\left( x_{t} \right)} \right\|^{2}}} \leq \frac{{\sum\limits_{t = 1}^{T}\left\lbrack {KM_{r}{\gamma_{t}}^{3}L\sigma^{2} + L{\gamma_{t}}^{2}\sigma^{2}} \right\rbrack} + \frac{2\left\lbrack {f\left( x_{1} \right) - f\left( x^{*} \right)} \right\rbrack}{KM_{r}}}{\sum\limits_{t = 1}^{T}\gamma_{t}}.
\end{equation}
\textbf{Proof} Given in Appendix.\\
Taking a close look at Theorem 1, we can properly choose the learning rate $\gamma_{t}$ as a constant value and obtain the following convergence rate.

\textbf{Corollary 2} Assume that Assumptions 1-5 hold. Set the learning rate $\gamma_{t}$ to be a constant $\gamma = \sqrt{\frac{f\left( x_{1} \right) - f\left( x^{*} \right)}{KM_{r}LT\sigma^{2}}}$, if the iteration number $T$ is bounded by: \\
\begin{equation}
    T \geq \frac{\left\lbrack {f\left( x_{1} \right) - f\left( x^{*} \right)} \right\rbrack KM_{r}}{9\sigma^{2}L},
\end{equation}
then the output of Algorithm 1 satisfies the following ergodic convergence rate:
\begin{equation}
    {\sum\limits_{t = 1}^{T}{\frac{1}{T}\left\| {\nabla f\left( x_{t} \right)} \right\|^{2}}} \leq 6\sigma\sqrt{\frac{\left\lbrack {f\left( x_{1} \right) - f\left( x^{*} \right)} \right\rbrack L}{TKM_{r}}} \leq O\left( \frac{1}{\sqrt{TKM_{r}}} \right).
\end{equation}
\textbf{Proof} Given in Appendix.\\
This corollary basically claims that when the iteration number $T \geq O\left( KM_{r} \right)$, the convergence rate of ABS-SGD achieves $O\left( \frac{1}{\sqrt{TKM_{r}}} \right)$. In homogeneous clusters, the batch size of each iteration is usually not too large, making $K$ close to $N$ and the convergence rate of ABS-SGD close to $O\left( \frac{1}{\sqrt{TNM_{r}}} \right)$. In heterogeneous clusters, since ABS-SGD can compute a larger batch size in an iteration and $K$ would be greater than $N$, ABS-SGD achieves a better convergence rate than in homogeneous clusters. However, since $T$ is bounded by $M_r$ in (10), the batch size is not the larger the better, as an excessively large batch size requires more iterations to achieve convergence.
\section{Experiments}
\subsection{Experimental settings}
\subsubsection{Experimental environment}
The experimental environment is a Linux server with 4 NVIDIA GeForce RTX 2080Ti GPUs, and the GPUs communicate through PCIE 3.0 x8. Each GPU can be regarded as a worker, so there are 4 workers in total. Due to the consistent configuration between GPUs and the absence of resource competition and other causes of dynamic heterogeneity, the experimental environment is a homogeneous cluster, and it is necessary to simulate heterogeneous clusters on this basis. Heterogeneous clusters are divided into clusters with only static heterogeneity, only dynamic heterogeneity, and both static and dynamic heterogeneity. The methods for simulating these three types of heterogeneous clusters are introduced below.

For clusters with only static heterogeneity, there are only differences in hardware performance, so the performance of each worker is relatively stable and would not change significantly in the short term. In view of the characteristics of static heterogeneity, we degrade worker performance by prolonging a fixed time in each iteration. Prolonging different fixed time can simulate workers with different hardware performance.

For clusters with only dynamic heterogeneity, e.g., cloud computing platform, there might be situations where multiple users and multiple jobs share the same node. Therefore, there is competition among jobs for resources, e.g., GPU, CPU, and memory. In view of the characteristics of dynamic heterogeneity, we prolong a random time to increase the iteration time in each iteration, so that the performance of workers would change dynamically. The ranges of the random time reflect the degree of dynamic heterogeneity.

For clusters with both static and dynamic heterogeneity, there are not only differences in hardware performance between workers, but also dynamic heterogeneity, e.g., multi-job resource competition. Therefore, we combine the simulation methods of the first two types of heterogeneous clusters. For static heterogeneity, it is simulated by prolonging a fixed time in each iteration; for dynamic heterogeneity, it is simulated by prolonging a random time in each iteration.
\subsubsection{Dataset and model}
We use CIFAR10 dataset to test the performance of each algorithm. CIFAR10 is a dataset for image classification, which contains a total of 60,000 pictures divided into 10 categories. The training set contains 50,000 pictures, and the test set contains 10,000 pictures. We train ResNet18 on CIFAR10 dataset, which has 11 million trainable parameters.
\subsubsection{Baselines and evaluation metric}
We conduct performance comparison with the following four commonly used baseline algorithms.
\begin{itemize}
\item BSP-SGD \cite{gerbessiotis1994direct}: BSP-SGD is the most commonly used distributed data parallel training algorithm, which requires all workers to synchronize gradients in each iteration.
\item DBS-SGD \cite{ye2020dbs}: On the basis of BSP-SGD, the batch size of DBS-SGD is dynamically adjusted based on performance prediction. The batch size of the next epoch is adjusted based on the performance of workers in the previous epoch.
\item ASP-SGD \cite{lian2015asynchronous}: In ASP-SGD, workers compute gradients based on a global model and update the global model asynchronously.
\item SSP-SGD \cite{ho2013more}: On the basis of ASP-SGD, SSP-SGD limits the difference in the iteration numbers between the fastest worker and the slowest worker to control the staleness of delayed gradients.
\end{itemize}
ABS-SGD and all the baseline algorithms are implemented based on the PyTorch deep learning framework. The communication parts of ABS-SGD, BSP-SGD, and DBS-SGD are implemented based on the AllReduce function provided by the PyTorch. The communication parts of ASP-SGD and SSP-SGD are implemented based on the Ray distributed framework.

The evaluation metric is the time required for the model to converge to a certain test accuracy. The convergence time of a model depends on the iteration time in each iteration and the iteration number required to achieve convergence, i.e., the hardware efficiency and statistical efficiency of an algorithm. Therefore, this evaluation metric can comprehensively reflect the hardware efficiency and statistical efficiency of an algorithm.

\subsubsection{Hyper-parameters}
The reference batch size of ABS-SGD is 32, and the batch size of each worker of baseline algorithms is 32. The compensation coefficient $\lambda$ of ABS-SGD is 0.5. The staleness threshold of SSP-SGD is 10. The learning rate is 0.01, and the total number of iterations is 6,200.
        \vspace{-0.8cm}
	\begin{figure}[H]
        \centering
	\subfigure[Only static heterogeneity]{
	\includegraphics[width=6cm,height=6cm]{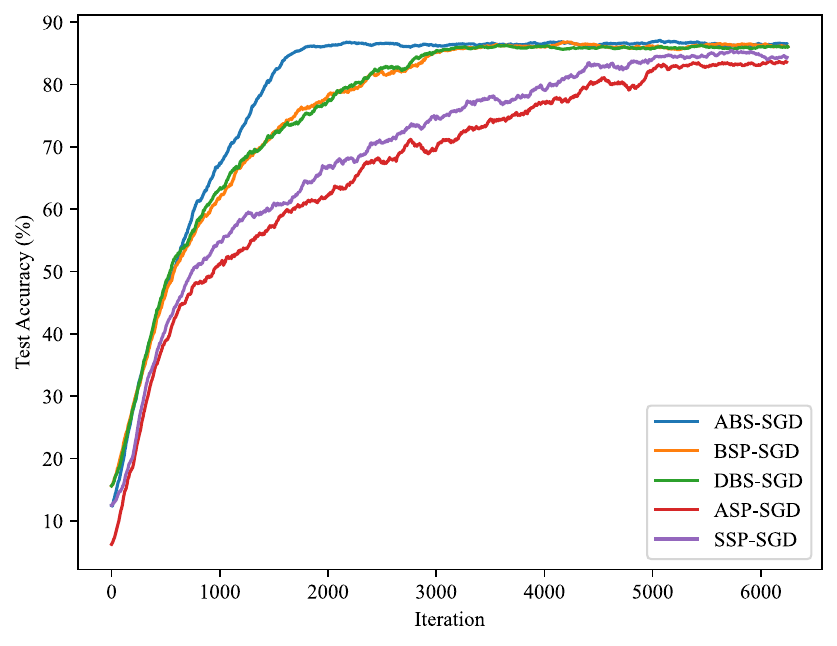} \label{Fig.6(b)}
}
	\hspace{2mm}
	\subfigure[Only dynamic heterogeneity]{
	\includegraphics[width=6cm,height=6cm]{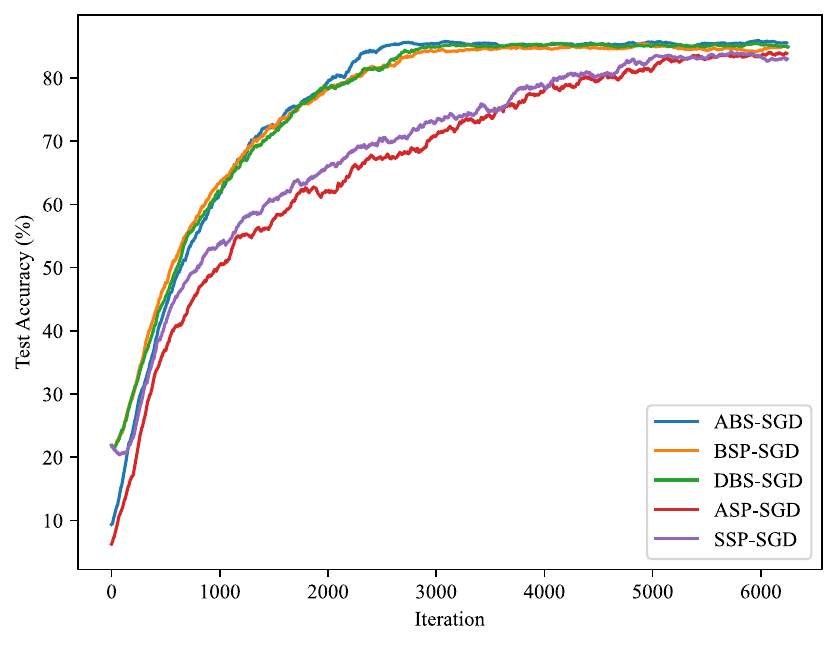} \label{Fig.6(b)}
}	
	\hspace{2mm}
	\subfigure[Both static and dynamic heterogeneity]{
		\includegraphics[width=6cm,height=6cm]{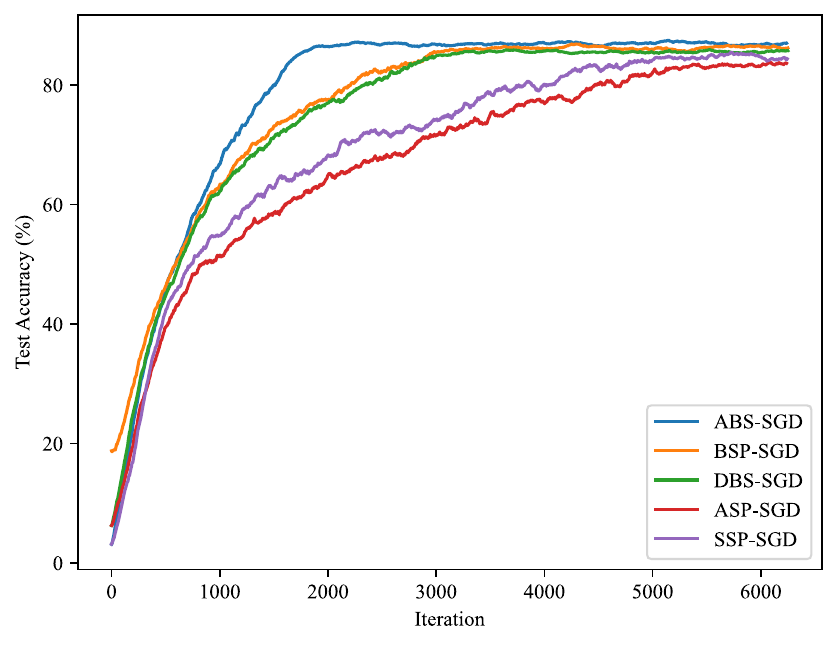} \label{Fig.6(b)}
	}
	
	\caption{Test accuracy for ResNet18 on CIFAR10 w.r.t iterations in three types of clusters.}
        \label{fig2}
	\end {figure}
 
        \vspace{-0.4cm}
	\begin{figure}[H]
        \centering
	\subfigure[Only static heterogeneity]{
	\includegraphics[width=6cm,height=6cm]{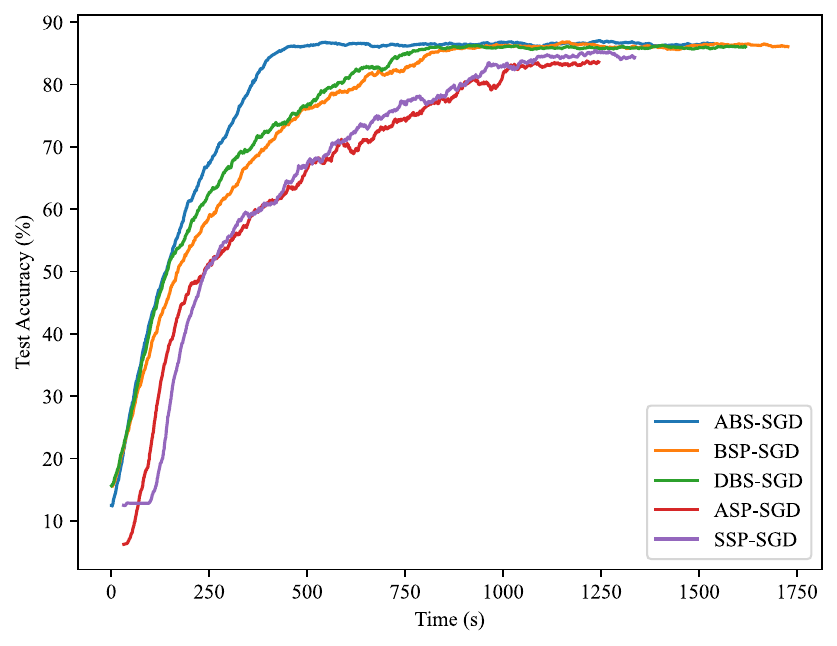} \label{Fig.6(b)}
}
	\hspace{2mm}
	\subfigure[Only dynamic heterogeneity]{
	\includegraphics[width=6cm,height=6cm]{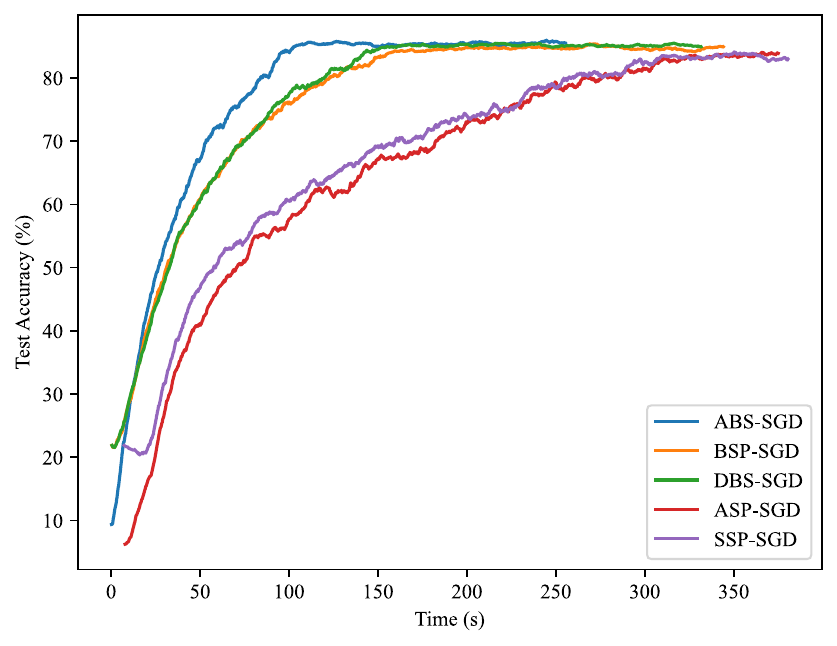} \label{Fig.6(b)}
}	
	\hspace{2mm}
	\subfigure[Both static and dynamic heterogeneity]{
		\includegraphics[width=6cm,height=6cm]{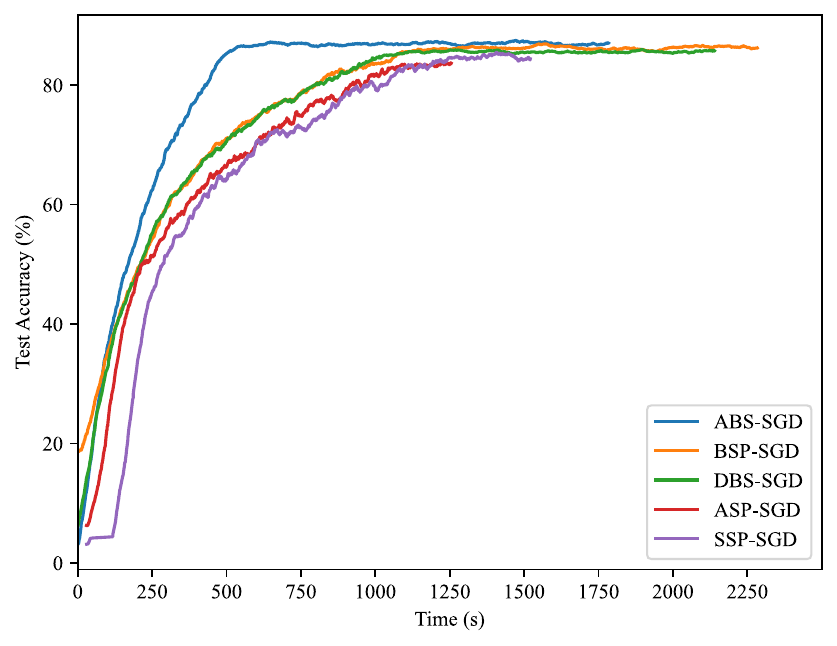} \label{Fig.6(b)}
	}
	
	\caption{Test accuracy for ResNet18 on CIFAR10 w.r.t time in three types of clusters.}
        \label{fig3}
	\end {figure}

\begin{table}[H]
  \caption{Convergence time (seconds) for all algorithms to achieve 80\% test accuracy in three types of clusters. The numbers in brackets indicate the speedup of ABS-SGD compared to other algorithms in the same cluster heterogeneity.}
  \label{table1}
  \centering
  \begin{tabular}{llllll}
    \toprule
    \makecell{Cluster heterogeneity} & \makecell{ABS-SGD} &  \makecell{BSP-SGD} &  \makecell{DBS-SGD} &  \makecell{ASP-SGD} &  \makecell{SSP-SGD}\\
    \midrule
    Only static & 450(1.00×)  & 636(1.41×) & 577(1.28×) & 900(2.00×) & 890(1.98×)     \\
    Only dynamic & 105(1.00×)  & 123(1.17×) & 120(1.14×) & 265(2.52×) & 254(2.42×)     \\
    Both static and dynamic & 535(1.00×)  & 798(1.49×) & 795(1.49×) & 916(1.71×) & 966(1.81×)     \\
    \bottomrule
  \end{tabular}
\end{table}

\subsection{Cluster with only static heterogeneity}
The cluster sets the iteration time to 1:2:3:4 for 4 workers. Therefore, we prolong the real iteration time of 4 workers in each iteration by 0\%, 100\%, 200\%, and 300\%, respectively. In order to ensure real-time performance, the real iteration time of each iteration will be recorded.

\textbf{Comparison of convergence rate} Figure \ref{fig2}(a) shows the test accuracy regarding iteration number when ResNet18 is trained in the cluster. Compared with other algorithms, ABS-SGD has higher statistical efficiency, i.e., requiring fewer iterations to converge, as ABS-SGD can utilize computational resources to compute larger size batches in each iteration. In terms of test accuracy, ABS-SGD has the same test accuracy as BSP-SGD and DBS-SGD, and has higher test accuracy than ASP-SGD and SSP-SGD, which is because the gradient delay of ABS-SGD is only one iteration, while ASP-SGD does not limit the gradient delay, and SSP-SGD limits gradient delay to 10.

\textbf{Comparison of convergence speed} Table \ref{table1} shows the convergence time for each algorithm to achieve 80\% test accuracy when ResNet18 is trained in the cluster. ABS-SGD has the fastest convergence speed, which is 1.41 times than that of BSP-SGD, 1.28 times than that of DBS-SGD, 2.00 times and 1.98 times than that of the two asynchronous algorithms ASP-SGD and SSP-SGD, respectively. Figure \ref{fig3}(a) shows the test accuracy regarding time during training with total 6,200 iterations. Compared with BSP-SGD and DBS-SGD, ABS-SGD takes less time to complete training, i.e., having higher hardware efficiency. This is because ABS-SGD can perform computation and communication in parallel to speed up iterations. Compared with ABS-SGD, ASP-SGD and SSP-SGD take less time to complete training, i.e., having higher hardware efficiency, but take longer to achieve convergence.

\subsection{Cluster with only dynamic heterogeneity}
The cluster assumes that the hardware performance of workers is consistent, but there are large performance fluctuations due to resource competition and other causes, so the range of random time is set between 0\% and 50\% of the real iteration time in each iteration. In order to ensure real-time performance, the real iteration time in each iteration will be recorded.

\textbf{Comparison of convergence rate} Figure \ref{fig2}(b) shows the test accuracy regarding iteration number when ResNet18 is trained in the cluster. ABS-SGD has the highest statistical efficiency, but it does not significantly exceed BSP-SGD and DBS-SGD. This is because the batch size of ABS-SGD in the cluster is not too large compared with that of BSP-SGD and DBS-SGD, as the performance gap in the cluster is not as large as that in the cluster with only static heterogeneity (the larger the performance gap between workers, the larger the batch size of ABS-SGD).

\textbf{Comparison of convergence speed} Table \ref{table1} shows the convergence time for each algorithm to achieve 80\% test accuracy when ResNet18 is trained in the cluster. ABS-SGD still has the fastest convergence speed, which is 1.17 times than that of BSP-SGD, 1.14 times than that of DBS-SGD, 2.52 times and 2.42 times than that of ASP-SGD and SSP-SGD, respectively. Compared with the experimental results in the cluster with only static heterogeneity, the convergence speed gap between ABS-SGD, BSP-SGD, and DBS-SGD is narrowing, which is also due to the smaller performance gap between workers. Figure \ref{fig3}(b) shows the test accuracy regarding time during training with total 6,200 iterations. Compared with the experimental results in the cluster with only static heterogeneity, the two asynchronous algorithms ASP-SGD and SSP-SGD are downgraded from the top two hardware efficiencies to the last two hardware efficiencies. This is because the performance gap in the cluster is not too large, and asynchronous algorithms have no advantage in communication over synchronous algorithms.

\subsection{Cluster with static and dynamic heterogeneity}
The heterogeneity of the cluster combines the configurations of the previous two clusters. For static heterogeneity, the real iteration time of 4 workers in each iteration will be prolonged by 0\%, 100\%, 200\%, and 300\%, respectively. For dynamic heterogeneity, the range of random time is set between 0\% and 50\% of the iteration time adjusted by static heterogeneity. Similarly, the real iteration time in each iteration will be recorded.

\textbf{Comparison of convergence rate} Figure \ref{fig2}(c) shows the test accuracy regarding iteration number when ResNet18 is trained in the cluster. Same as the experimental results in the cluster with only static heterogeneity, ABS-SGD has higher statistical efficiency than other algorithms. This is because when both dynamic heterogeneity and static heterogeneity exist, the performance gap between workers would be large, so that ABS-SGD can use computational resources to compute larger size batches. In terms of the test accuracy, similar to the results in the previous two clusters, the test accuracy of ABS-SGD is the same as that of BSP-SGD and DBS-SGD, and ABS-SGD has a higher test accuracy than ASP-SGD and SSP-SGD.

\textbf{Comparison of convergence speed} Table \ref{table1} shows the convergence time for each algorithm to achieve 80\% test accuracy when ResNet18 is trained in the cluster. ABS-SGD still has the fastest convergence speed, which is 1.49 times than that of BSP-SGD and DBS-SGD, 1.71 times and 1.81 times than that of ASP-SGD and SSP-SGD, respectively. Among the three clusters, the convergence speed improvement of ABS-SGD over BSP-SGD and DBS-SGD is largest in the cluster with static and dynamic heterogeneity. This result indicates that ABS-SGD can make full use of computational resources in complicated heterogeneous environments, adapting the batch size of each worker to its performance, thereby speeding up model convergence. Figure \ref{fig3}(c) shows the test accuracy regarding time during training with total 6,200 iterations, which is similar to the experimental results in the cluster with only static heterogeneity. Compared with BSP-SGD and DBS-SGD, ABS-SGD has higher hardware efficiency, but not as good as ASP-SGD and SSP-SGD.

\section{Conclusions and Future Work}
In this paper, we propose a delayed synchronous SGD algorithm with adaptive batch size (ABS-SGD) for heterogeneous GPU clusters. In ABS-SGD, each worker performs the gradient computation of the next batch in parallel with global synchronization for delayed gradients in each iteration. The batch size is not specified in advance and the gradient computation will continue until the end of global synchronization, so as to fully utilize computational resources. ABS-SGD is also theoretically proven to have a linear iteration speedup with respect to the total computational resources. In addition, experimental results demonstrate the efficiency of ABS-SGD, increasing the convergence speed by 1.30× on average compared with the best baseline algorithm.

In the future, we will extend our work in the following directions. Firstly, since ABS-SGD just uses the plain SGD algorithm to update model parameters, we will combine the heterogeneous-aware delayed synchronous parallel mechanism with more sophisticated algorithms, e.g., LARS or Adam, to improve the algorithm. Secondly, we will introduce the idea of Local SGD, i.e., allowing workers to perform multiple local iterations according to their performance, which can further improve the convergence speed by decreasing communication frequency.

\bibliographystyle{unsrt}
\bibliography{ref.bib}

\begin{thebibliography}{10}

\bibitem{goyal2017accurate}
Priya Goyal, Piotr Doll{\'a}r, Ross Girshick, Pieter Noordhuis, Lukasz
  Wesolowski, Aapo Kyrola, Andrew Tulloch, Yangqing Jia, and Kaiming He.
\newblock {Accurate, Large Minibatch SGD: Training ImageNet in 1 Hour}.
\newblock {\em arXiv preprint arXiv:1706.02677}, 2017.

\bibitem{jia2018highly}
Xianyan Jia, Shutao Song, Wei He, Yangzihao Wang, Haidong Rong, Feihu Zhou,
  Liqiang Xie, Zhenyu Guo, Yuanzhou Yang, Liwei Yu, et~al.
\newblock {Highly Scalable Deep Learning Training System with Mixed-Precision:
  Training ImageNet in Four Minutes}.
\newblock {\em arXiv preprint arXiv:1807.11205}, 2018.

\bibitem{hazelwood2018applied}
Kim Hazelwood, Sarah Bird, David Brooks, Soumith Chintala, Utku Diril, Dmytro
  Dzhulgakov, Mohamed Fawzy, Bill Jia, Yangqing Jia, Aditya Kalro, et~al.
\newblock {Applied Machine Learning at Facebook: A Datacenter Infrastructure
  Perspective}.
\newblock In {\em IEEE International Symposium on High Performance Computer
  Architecture}, 2018.

\bibitem{jeon2018multi}
Myeongjae Jeon, Shivaram Venkataraman, Junjie Qian, Amar Phanishayee, Wencong
  Xiao, and Fan Yang.
\newblock {Multi-tenant GPU Clusters for Deep Learning Workloads: Analysis and
  Implications}.
\newblock {\em Technical report, Microsoft Research}, 2018.

\bibitem{gerbessiotis1994direct}
Alexandros~V Gerbessiotis and Leslie~G Valiant.
\newblock {Direct Bulk-Synchronous Parallel Algorithms}.
\newblock {\em Journal of Parallel and Distributed Computing}, 22(2):251--267,
  1994.

\bibitem{cipar2013solving}
James Cipar, Qirong Ho, Jin~Kyu Kim, Seunghak Lee, Gregory~R Ganger, Garth
  Gibson, Kimberly Keeton, and Eric P~Xing.
\newblock {Solving the Straggler Problem with Bounded Staleness}.
\newblock In {\em USENIX Workshop on Hot Topics in Operating Systems}, 2013.

\bibitem{zinkevich2010parallelized}
Martin Zinkevich, Markus Weimer, Lihong Li, and Alex Smola.
\newblock {Parallelized Stochastic Gradient Descent}.
\newblock In {\em Advances in Neural Information Processing Systems}, 2010.

\bibitem{lian2015asynchronous}
Xiangru Lian, Yijun Huang, Yuncheng Li, and Ji~Liu.
\newblock {Asynchronous Parallel Stochastic Gradient for Nonconvex
  Optimization}.
\newblock In {\em Advances in Neural Information Processing Systems}, 2015.

\bibitem{ho2013more}
Qirong Ho, James Cipar, Henggang Cui, Seunghak Lee, Jin~Kyu Kim, Phillip~B
  Gibbons, Garth~A Gibson, Greg Ganger, and Eric~P Xing.
\newblock {More Effective Distributed ML via a Stale Synchronous Parallel
  Parameter Server}.
\newblock In {\em Advances in Neural Information Processing Systems}, 2013.

\bibitem{ananthanarayanan2013effective}
Ganesh Ananthanarayanan, Ali Ghodsi, Scott Shenker, and Ion Stoica.
\newblock {Effective Straggler Mitigation: Attack of the Clones.}
\newblock In {\em USENIX Symposium on Networked Systems Design and
  Implementation}, 2013.

\bibitem{zaharia2008improving}
Matei Zaharia, Andy Konwinski, Anthony~D Joseph, Randy~H Katz, and Ion Stoica.
\newblock {Improving MapReduce Performance in Heterogeneous Environments}.
\newblock In {\em USENIX Symposium on Operating Systems Design and
  Implementation}, 2008.

\bibitem{chen2016revisiting}
Jianmin Chen, Xinghao Pan, Rajat Monga, Samy Bengio, and Rafal Jozefowicz.
\newblock {Revisiting Distributed Synchronous SGD}.
\newblock {\em arXiv preprint arXiv:1604.00981}, 2016.

\bibitem{li2020taming}
Shigang Li, Tal Ben-Nun, Salvatore~Di Girolamo, Dan Alistarh, and Torsten
  Hoefler.
\newblock {Taming Unbalanced Training Workloads in Deep Learning with Partial
  Collective Operations}.
\newblock In {\em ACM SIGPLAN Symposium on Principles and Practice of Parallel
  Programming}, 2020.

\bibitem{jiang2017heterogeneity}
Jiawei Jiang, Bin Cui, Ce~Zhang, and Lele Yu.
\newblock {Heterogeneity-aware Distributed Parameter Servers}.
\newblock In {\em ACM International Conference on Management of Data}, 2017.

\bibitem{zhao2019dynamic}
Xing Zhao, Aijun An, Junfeng Liu, and Bao~Xin Chen.
\newblock {Dynamic Stale Synchronous Parallel Distributed Training for Deep
  Learning}.
\newblock In {\em IEEE International Conference on Distributed Computing
  Systems}, 2019.

\bibitem{stich2018local}
Sebastian~U Stich.
\newblock {Local SGD Converges Fast and Communicates Little}.
\newblock {\em arXiv preprint arXiv:1805.09767}, 2018.

\bibitem{ye2020dbs}
Qing Ye, Yuhao Zhou, Mingjia Shi, Yanan Sun, and Jiancheng Lv.
\newblock {DBS: Dynamic Batch Size for Distributed Deep Neural Network
  Training}.
\newblock {\em arXiv preprint arXiv:2007.11831}, 2020.

\bibitem{tyagi2020taming}
Sahil Tyagi and Prateek Sharma.
\newblock {Taming Resource Heterogeneity in Distributed ML Training with
  Dynamic Batching}.
\newblock In {\em IEEE International Conference on Autonomic Computing and
  Self-Organizing Systems}, 2020.

\bibitem{chen2020semi}
Chen Chen, Qizhen Weng, Wei Wang, Baochun Li, and Bo~Li.
\newblock {Semi-Dynamic Load Balancing: Efficient Distributed Learning in
  Non-Dedicated Environments}.
\newblock In {\em ACM Symposium on Cloud Computing}, 2020.

\bibitem{stripelis2021semi}
Dimitris Stripelis and Jos{\'e}~Luis Ambite.
\newblock {Semi-Synchronous Federated Learning}.
\newblock {\em arXiv preprint arXiv:2102.02849}, 2021.

\bibitem{cao2021sap}
Jing Cao, Zongwei Zhu, and Xuehai Zhou.
\newblock {SAP-SGD: Accelerating Distributed Parallel Training with High
  Communication Efficiency on Heterogeneous Clusters}.
\newblock In {\em IEEE International Conference on Cluster Computing}, 2021.

\bibitem{jiang2019novel}
Wenbin Jiang, Geyan Ye, Laurence~T Yang, Jian Zhu, Yang Ma, Xia Xie, and Hai
  Jin.
\newblock {A Novel Stochastic Gradient Descent Algorithm Based on Grouping over
  Heterogeneous Cluster Systems for Distributed Deep Learning}.
\newblock In {\em IEEE/ACM International Symposium on Cluster, Cloud and Grid
  Computing}, 2019.

\bibitem{sun2021gssp}
Haifeng Sun, Zhiyi Gui, Song Guo, Qi~Qi, Jingyu Wang, and Jianxin Liao.
\newblock {GSSP: Eliminating Stragglers through Grouping Synchronous for
  Distributed Deep Learning in Heterogeneous Cluster}.
\newblock {\em IEEE Transactions on Cloud Computing}, 10(4):2637--2648, 2021.

\bibitem{zheng2017asynchronous}
Shuxin Zheng, Qi~Meng, Taifeng Wang, Wei Chen, Nenghai Yu, Zhi-Ming Ma, and
  Tie-Yan Liu.
\newblock {Asynchronous Stochastic Gradient Descent with Delay Compensation}.
\newblock In {\em International Conference on Machine Learning}, 2017.

\end{thebibliography}

\newpage
\setcounter{equation}{0}
\appendix

\section{Appendix}

\textbf{Proofs to Theorem 1} \\
From the Lipschitzisan gradient Assumption 1, we have
\begin{align}
f\left( x_{t + 1} \right) & \leq f\left( x_{t} \right) + \left\langle {\nabla f\left( x_{t} \right),x_{t + 1} - x_{t}} \right\rangle + \frac{L}{2}\left\| {x_{t + 1} - x_{t}} \right\|^{2}  \notag \\
& \leq - \gamma_{t}\left\langle {\nabla f\left( x_{t} \right),{\sum\limits_{m = 1}^{M_{t}}{g\left( {x_{t - 1};\xi_{t - 1,m}} \right)}}} \right\rangle + \frac{L{\gamma_{t}}^{2}}{2}\left\| {\sum\limits_{m = 1}^{M_{t}}{g\left( {x_{t - 1};\xi_{t - 1,m}} \right)}} \right\|^{2},
\end{align}
where the second inequality is due to $x_{t + 1} - x_{t} = - \gamma_{t}{\sum\limits_{m = 1}^{M_{t}}{G\left( x_{t - 1};\xi_{t - 1,m} \right)}}$. \\
We next take expectation respect to $\xi$ on both sides of (1). We have
\begin{equation}
    f\left( x_{t + 1} \right) - f\left( x_{t} \right) \leq - \gamma_{t}M_{t}\left\langle {\nabla f\left( x_{t} \right),\nabla f\left( x_{t - 1} \right)} \right\rangle + \frac{L{\gamma_{t}}^{2}}{2}E\left\lbrack \left\| {\sum\limits_{m = 1}^{M_{t}}{g\left( {x_{t - 1};\xi_{t - 1,m}} \right)}} \right\|^{2} \right\rbrack.
\end{equation}
From the fact $\left\langle {a,b} \right\rangle = \frac{1}{2}\left( \left\| a \right\|^{2} + \left\| b \right\|^{2} - \left\| {a - b} \right\|^{2} \right)$, we have
\begin{align}
f\left( x_{t + 1} \right) - f\left( x_{t} \right) \leq & - \frac{\gamma_{t}M_{t}}{2}\left\lbrack {\left\| {\nabla f\left( x_{t} \right)} \right\|^{2} + \left\| {\nabla f\left( x_{t - 1} \right)} \right\|^{2} - \underset{T1}{\underbrace{\left\| {\nabla f\left( x_{t} \right) - \nabla f\left( x_{t - 1} \right)} \right\|^{2}}}} \right\rbrack \notag \\
& + \frac{L{\gamma_{t}}^{2}}{2}\underset{T2}{\underbrace{E\left\lbrack \left\| {\sum\limits_{m = 1}^{M_{t}}{g\left( {x_{t - 1};\xi_{t - 1,m}} \right)}} \right\|^{2} \right\rbrack}}.
\end{align}
Next we estimate the upper bound of $T1$ and $T2$. For $T2$ we have
\begin{align}
    T2 & = E\left\lbrack \left\| {\sum\limits_{m = 1}^{M_{t}}{g\left( {x_{t - 1};\xi_{t - 1,m}} \right)}} \right\|^{2} \right\rbrack \notag \\
    & = E\left\lbrack \left\| {{\sum\limits_{m = 1}^{M_{t}}\left( {g\left( {x_{t - 1};\xi_{t - 1,m}} \right) - \nabla f\left( x_{t - 1} \right)} \right)} + {\sum\limits_{m = 1}^{M_{t}}{\nabla f\left( x_{t - 1} \right)}}} \right\|^{2} \right\rbrack \notag \\
    \begin{split}
    & = E\left\lbrack \left\| {\sum\limits_{m = 1}^{M_{t}}\left( {g\left( {x_{t - 1};\xi_{t - 1,m}} \right) - \nabla f\left( x_{t - 1} \right)} \right)} \right\|^{2} + \left\| {\sum\limits_{m = 1}^{M_{t}}{\nabla f\left( x_{t - 1} \right)}} \right\|^{2} \right. \notag \\ 
    &\qquad \left. + 2\left\langle {{\sum\limits_{m = 1}^{M_{t}}\left( {g\left( {x_{t - 1};\xi_{t - 1,m}} \right) - \nabla f\left( x_{t - 1} \right)} \right)},{\sum\limits_{m = 1}^{M_{t}}{\nabla f\left( x_{t - 1} \right)}}} \right\rangle \right\rbrack \notag \\
    \end{split}
    \\
    & = E\left\lbrack {\left\| {\sum\limits_{m = 1}^{M_{t}}\left( {g\left( {x_{t - 1};\xi_{t - 1,m}} \right) - \nabla f\left( x_{t - 1} \right)} \right)} \right\|^{2} + \left\| {\sum\limits_{m = 1}^{M_{t}}{\nabla f\left( x_{t - 1} \right)}} \right\|^{2}} \right\rbrack \notag \\
    \begin{split}
    & = E\left\lbrack {\sum\limits_{m = 1}^{M_{t}}\left\| \left( {g\left( {x_{t - 1};\xi_{t - 1,m}} \right) - \nabla f\left( x_{t - 1} \right)} \right) \right\|^{2}} \right\rbrack + \left\| {\sum\limits_{m = 1}^{M_{t}}{\nabla f\left( x_{t - 1} \right)}} \right\|^{2} \\
    & \qquad + 2{\sum\limits_{1 \leq m < m^{'} \leq M_{t}}\left\langle {g\left( {x_{t - 1};\xi_{t - 1,m}} \right) - \nabla f\left( x_{t - 1} \right),g\left( {x_{t - 1};\xi_{t - 1,m^{'}}} \right) - \nabla f\left( x_{t - 1} \right)} \right\rangle} \notag \\
    \end{split}
    \\
    & \leq M_{t}\sigma^{2} + {M_{t}}^{2}\left\| {\nabla f\left( x_{t - 1} \right)} \right\|^{2},
\end{align}
where the fourth equality is due to 
\begin{align}
& E\left\lbrack \left\langle {{\sum\limits_{m = 1}^{M_{t}}\left( {g\left( {x_{t - 1};\xi_{t - 1,m}} \right) - \nabla f\left( x_{t - 1} \right)} \right)},{\sum\limits_{m = 1}^{M_{t}}{\nabla f\left( x_{t - 1} \right)}}} \right\rangle \right\rbrack \notag \\
& = \left\langle {{\sum\limits_{m = 1}^{M_{t}}{E\left\lbrack \left( {g\left( {x_{t - 1};\xi_{t - 1,m}} \right) - \nabla f\left( x_{t - 1} \right)} \right) \right\rbrack}},{\sum\limits_{m = 1}^{M_{t}}{\nabla f\left( x_{t - 1} \right)}}} \right\rangle \notag \\
& = 0,
\end{align}
and the last inequality is due to the Assumption 3 and
\begin{align}
    & E\left\lbrack {\sum\limits_{1 \leq m < m^{'} \leq M_{t}}\left\langle {g\left( {x_{t - 1};\xi_{t - 1,m}} \right) - \nabla f\left( x_{t - 1} \right),g\left( {x_{t - 1};\xi_{t - 1,m^{'}}} \right) - \nabla f\left( x_{t - 1} \right)} \right\rangle} \right\rbrack \notag \\
    & = {\sum\limits_{1 \leq m < m^{'} \leq M_{t}}\left\langle {g\left( {x_{t - 1};\xi_{t - 1,m}} \right) - \nabla f\left( x_{t - 1} \right),E\left\lbrack {g\left( {x_{t - 1};\xi_{t - 1,m^{'}}} \right) - \nabla f\left( x_{t - 1} \right)} \right\rbrack} \right\rangle} \notag \\
    & = 0.
\end{align}
For $T1$ we have
\begin{align}
T1 & = \left\| {\nabla f\left( x_{t} \right) - \nabla f\left( x_{t - 1} \right)} \right\|^{2} \notag \\
& \leq L\left\| {x_{t} - x_{t - 1}} \right\|^{2} \notag \\
& \leq {\gamma_{t}}^{2}L\left\| {\sum\limits_{m = 1}^{M_{t}}{G\left( {x_{t - 1};\xi_{t - 1,m}} \right)}} \right\|^{2} \notag \\
& \leq {\gamma_{t}}^{2}LM_{t}\sigma^{2} + {\gamma_{t}}^{2}L{M_{t}}^{2}\left\| {\nabla f\left( x_{t - 1} \right)} \right\|^{2},
\end{align}
where the first inequality is due to the Assumption 1 and the third inequality is due to (4).
Applying the upper bounds for $T1$ in (7) and $T2$ in (4), we obtain
\begin{align}
    f\left( x_{t + 1} \right) - f\left( x_{t} \right) & \leq - \frac{\gamma_{t}M_{t}}{2}\left\lbrack \left\| {\nabla f\left( x_{t} \right)} \right\|^{2} + \left\| {\nabla f\left( x_{t - 1} \right)} \right\|^{2} \right. \notag \\ 
    \begin{split}
        & \qquad \left. - \left( {{\gamma_{t}}^{2}LM_{t}\sigma^{2} + {\gamma_{t}}^{2}L{M_{t}}^{2}\left\| {\nabla f\left( x_{t - 1} \right)} \right\|^{2}} \right) \right\rbrack \notag \\
        &\qquad + \frac{L{\gamma_{t}}^{2}}{2}\left( {M_{t}\sigma^{2} + {M_{t}}^{2}\left\| {\nabla f\left( x_{t - 1} \right)} \right\|^{2}} \right) \notag \\
    \end{split}
    \\
    & \leq - \frac{\gamma_{t}M_{t}}{2}\left\| {\nabla f\left( x_{t} \right)} \right\|^{2} + \frac{{M_{t}}^{2}{\gamma_{t}}^{3}L\sigma^{2} + L{\gamma_{t}}^{2}M_{t}\sigma^{2}}{2} \notag \\
    & \qquad + \frac{{\gamma_{t}}^{3}L{M_{t}}^{3} + L{\gamma_{t}}^{2}{M_{t}}^{2} - \gamma_{t}M_{t}}{2}\left\| {\nabla f\left( x_{t - 1} \right)} \right\|^{2}.
\end{align}
Assume ${\gamma_{t}}^{3}L{M_{t}}^{3} + L{\gamma_{t}}^{2}{M_{t}}^{2} - \gamma_{t}M_{t} \leq 0$, we have ${\gamma_{t}}^{2}L{M_{t}}^{2} + L\gamma_{t}M_{t} \leq 1$ and then we have
\begin{align}
    f\left( x_{t + 1} \right) - f\left( x_{t} \right) \leq - \frac{\gamma_{t}M_{t}}{2}\left\| {\nabla f\left( x_{t} \right)} \right\|^{2} + \frac{{M_{t}}^{2}{\gamma_{t}}^{3}L\sigma^{2} + L{\gamma_{t}}^{2}M_{t}\sigma^{2}}{2}.
\end{align}
Summarizing (9) from $t=1$ to $t=T$, we have
\begin{align}
    f\left( x_{T + 1} \right) - f\left( x_{1} \right) \leq {\sum\limits_{t = 1}^{T}{- \frac{\gamma_{t}M_{t}}{2}\left\| {\nabla f\left( x_{t} \right)} \right\|^{2}}} + {\sum\limits_{t = 1}^{T}\left\lbrack {\frac{{M_{t}}^{2}{\gamma_{t}}^{3}L\sigma^{2}}{2} + \frac{L{\gamma_{t}}^{2}M_{t}\sigma^{2}}{2}} \right\rbrack}.
\end{align}
From the Bounded Delay Assumption 5, we have
\begin{align}
    f\left( x_{T + 1} \right) - f\left( x_{1} \right) \leq {\sum\limits_{t = 1}^{T}{- \frac{\gamma_{t}KM_{r}}{2}\left\| {\nabla f\left( x_{t} \right)} \right\|^{2}}} + {\sum\limits_{t = 1}^{T}\left\lbrack {\frac{K^{2}{M_{r}}^{2}{\gamma_{t}}^{3}L\sigma^{2}}{2} + \frac{L{\gamma_{t}}^{2}KM_{r}\sigma^{2}}{2}} \right\rbrack}.
\end{align}
Simplify (11), we have 
\begin{align}
    {\sum\limits_{t = 1}^{T}{\frac{\gamma_{t}KM_{r}}{2}\left\| {\nabla f\left( x_{t} \right)} \right\|^{2}}} & \leq {\sum\limits_{t = 1}^{T}\left\lbrack {\frac{{KM_{r}}^{2}{\gamma_{t}}^{3}L\sigma^{2}}{2} + \frac{L{\gamma_{t}}^{2}KM_{r}\sigma^{2}}{2}} \right\rbrack} + f\left( x_{1} \right) - f\left( x_{T + 1} \right) \notag \\
    {\sum\limits_{t = 1}^{T}{\gamma_{t}\left\| {\nabla f\left( x_{t} \right)} \right\|^{2}}} & \leq {\sum\limits_{t = 1}^{T}\left\lbrack {KM_{r}{\gamma_{k}}^{3}L\sigma^{2} + L{\gamma_{k}}^{2}\sigma^{2}} \right\rbrack} + \frac{2\left\lbrack {f\left( x_{1} \right) - f\left( x_{T + 1} \right)} \right\rbrack}{KM_{r}}.
\end{align}
Divide (12) by $\frac{1}{\sum\limits_{t = 1}^{T}\gamma_{t}}$, we have
\begin{align}
    \frac{1}{\sum\limits_{t = 1}^{T}\gamma_{t}}{\sum\limits_{t = 1}^{T}{\gamma_{t}\left\| {\nabla f\left( x_{t} \right)} \right\|^{2}}} \leq \frac{{\sum\limits_{t = 1}^{T}\left\lbrack {KM_{r}{\gamma_{t}}^{3}L\sigma^{2} + L{\gamma_{t}}^{2}\sigma^{2}} \right\rbrack} + \frac{2\left\lbrack {f\left( x_{1} \right) - f\left( x_{T + 1} \right)} \right\rbrack}{KM_{r}}}{\sum\limits_{t = 1}^{T}\gamma_{t}}.
\end{align}
Note that $x^{*}$ is the global optimization point. Thus we have
\begin{align}
    \frac{1}{\sum\limits_{t = 1}^{T}\gamma_{t}}{\sum\limits_{t = 1}^{T}{\gamma_{t}\left\| {\nabla f\left( x_{t} \right)} \right\|^{2}}} \leq \frac{{\sum\limits_{t = 1}^{T}\left\lbrack {KM_{r}{\gamma_{t}}^{3}L\sigma^{2} + L{\gamma_{t}}^{2}\sigma^{2}} \right\rbrack} + \frac{2\left\lbrack {f\left( x_{1} \right) - f\left( x^{*} \right)} \right\rbrack}{KM_{r}}}{\sum\limits_{t = 1}^{T}\gamma_{t}}.
\end{align}
It completes the proof.\\ \\
\textbf{Proofs to Corollary 2} \\
From (14) and assume $\gamma_{k} = \gamma$, we have
\begin{align}
    {\sum\limits_{t = 1}^{T}{\frac{1}{T}\left\| {\nabla f\left( x_{t} \right)} \right\|^{2}}} \leq \frac{T\sigma^{2}L\gamma^{2}\left\lbrack {K^{2}{M_{r}}^{2}\gamma + KM_{r}} \right\rbrack + 2\left\lbrack {f\left( x_{1} \right) - f\left( x^{*} \right)} \right\rbrack}{TKM_{r}\gamma}.
\end{align}
Simplify (15), we have
\begin{align}
    {\sum\limits_{t = 1}^{T}{\frac{1}{T}\left\| {\nabla f\left( x_{t} \right)} \right\|^{2}}} \leq \frac{2\left\lbrack {f\left( x_{1} \right) - f\left( x^{*} \right)} \right\rbrack}{TKM_{r}\gamma} + \sigma^{2}L\gamma + KM_{r}\sigma^{2}L\gamma^{2}.
\end{align}
Make the first term and the second term on the right side of (16) have the same order, i.e., $\gamma = \sqrt{\frac{f\left( x_{1} \right) - f\left( x^{*} \right)}{KM_{r}LT\sigma^{2}}}$, we have
\begin{align}
    {\sum\limits_{t = 1}^{T}{\frac{1}{T}\left\| {\nabla f\left( x_{t} \right)} \right\|^{2}}} & \leq \frac{2\left\lbrack {f\left( x_{1} \right) - f\left( x^{*} \right)} \right\rbrack}{TKM_{r}}\sqrt{\frac{KM_{r}LT\sigma^{2}}{f\left( x_{1} \right) - f\left( x^{*} \right)}} \notag \\
    & \qquad + \sigma^{2}L\sqrt{\frac{f\left( x_{1} \right) - f\left( x^{*} \right)}{KM_{r}LT\sigma^{2}}} + \frac{f\left( x_{1} \right) - f\left( x^{*} \right)}{T}.
\end{align}
Simplify (17), we have
\begin{align}
    {\sum\limits_{t = 1}^{T}{\frac{1}{T}\left\| {\nabla f\left( x_{t} \right)} \right\|^{2}}} \leq 3\sigma\sqrt{\frac{\left\lbrack {f\left( x_{1} \right) - f\left( x^{*} \right)} \right\rbrack L}{KM_{r}}} + \frac{f\left( x_{1} \right) - f\left( x^{*} \right)}{T}.
\end{align}
Then make the order of the first term on the right side of (18) larger than the order of the second term, we have
\begin{align}
    {\sum\limits_{t = 1}^{T}{\frac{1}{T}\left\| {\nabla f\left( x_{t} \right)} \right\|^{2}}} \leq 6\sigma\sqrt{\frac{\left\lbrack {f\left( x_{1} \right) - f\left( x^{*} \right)} \right\rbrack L}{TKM_{r}}} \leq O\left( \frac{1}{\sqrt{TKM_{r}}} \right).
\end{align}
It completes the proof.
\end{document}